\pdfoutput=1

\documentclass[11pt]{article}
\usepackage{float}
\usepackage{url}
\usepackage{graphicx}
\usepackage{booktabs}
\usepackage{multirow} 
\usepackage{footnote}
\usepackage{arydshln}
\usepackage{tablefootnote}
\usepackage{booktabs}  
\usepackage{threeparttable} 
\usepackage{tabularx}
\usepackage[table]{xcolor}
\usepackage[]{EACL2023}
\usepackage{array}
\usepackage{makecell}
\usepackage{times}
\usepackage{latexsym}

\usepackage[T1]{fontenc}

\usepackage[utf8]{inputenc}

\usepackage{microtype}

\title{Improving User Controlled Table-To-Text Generation Robustness}

\author{Hanxu Hu\textsuperscript{1 },  Yunqing Liu\textsuperscript{2}, Zhongyi Yu\textsuperscript{1} and Laura Perez-Beltrachini \textsuperscript{1}  \\
         \textsuperscript{1} School of Informatics, University of Edinburgh, United Kingdom \\ \textsuperscript{2} The Hong Kong Polytechnic University, HongKong\\ \texttt{\{huhanxu1233,lyq6175215241,zhongyics\}@gmail.com} \\
         \texttt{lperez@exseed.ed.ac.uk}}

\begin{document}
\maketitle
\begin{abstract}

In this work we study user controlled table-to-text generation 
where users explore the content in a table by selecting cells
and reading a natural language description
thereof automatically produce by a natural language generator.
Such generation models usually learn from carefully selected cell
combinations (clean cell selections); however, in practice
users may select unexpected, redundant, or incoherent cell 
combinations (noisy cell selections). 
In experiments, we find that models perform well on test sets coming 
from the same distribution as the train data but their performance drops
when evaluated on realistic noisy user inputs.
We propose a fine-tuning regime with additional user-simulated 
noisy cell selections.
Models fine-tuned with the proposed regime gain 4.85 BLEU points
on user noisy test cases and 1.4 on clean test cases; 
and achieve comparable
state-of-the-art performance on the ToTTo dataset.\footnote{Our code is available at \\ \href{https://github.com/hanxuhu/controllT2Trobust}{https://github.com/hanxuhu/controllT2Trobust}}

\end{abstract}

\section{Introduction}

\begin{figure}[tb]
\begin{center}
\centerline{\includegraphics[width=\columnwidth]{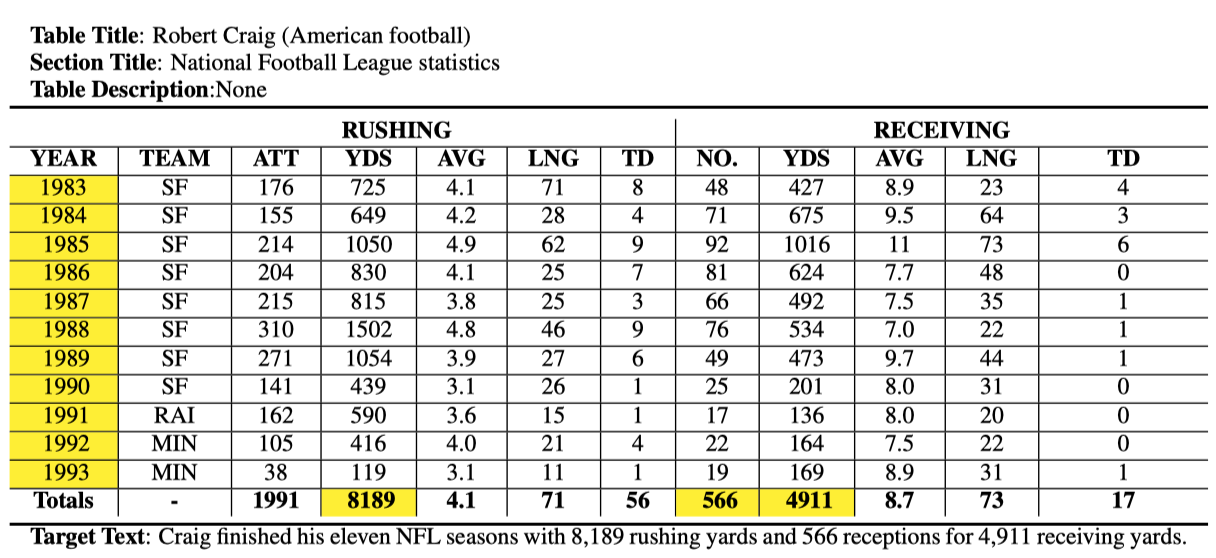}}
\caption{An example in the ToTTo dataset. The figure is retrieved from \cite{parikh2020totto}. The cells coloured in yellow are the highlight cells.}
\label{fig:totto_example}
\end{center}
\vskip -10mm
\end{figure}

The goal of table-to-text generation is to provide the user with a 
description of the most relevant content in a given table \cite{lebret-etal-2016neural,wiseman-etal-2018-learning,perez-lapata2018,puduppully2019data}. 
Recently, \citet{parikh2020totto} proposed a controlled table-to-text
generation task where the goal is to automatically create a description
for a determined subset of the table, namely the highlighted table cells.
The main focus on \citeauthor{parikh2020totto}'s \citeyear{parikh2020totto}
work is to assess the performance of neural text generators in a more 
controlled setting, i.e., when given an input table with explicit instructions 
(i.e., highlights) on what should be expressed in the output description. 
In this work, we view this task in the context of a natural language
interface, as a \emph{user controlled table-to-text}
generation task, where users provide those highlights interactively by 
exploring the content of a given table and study these user interactions.
Figure~\ref{fig:totto_example} illustrates the case where a user 
selects some cells (highlighted in yellow) and the generator 
provides a description thereof (shown below the table).

A crucial aspect of usability assessment for a generator in 
this interactive table-to-text task is robustness.
In a recent study by \cite{gem-sets-2021}, it has been shown that
neural generation models fail to maintain their in distribution
performance when confronted with realistic scenarios at test time 
such as typos in the input text. In the case of 
user controlled table-to-text generation, users may introduce noise 
when exploring the table content and select cell combinations that 
turn out to be  unexpected, redundant, or incoherent.
For example, in Figure~\ref{fig:totto_example}, when the user wants 
to express "eleven seasons", they might miss one year or highlight 
the header cell.
They may also select unrelated headers, for instance adding the header
"LNG" to the current selection. Existing controlled table-to-text
generation models 
\cite{parikh2020totto,su2021plan,kalerastogi-2020-text} 
are trained on carefully selected cell combinations (\textbf{clean} 
cell highlights) from the ToTTo dataset \cite{parikh2020totto}. 
We argue that these models will not generalize well 
in practice with user \textbf{noisy} highlights. 
No previous work has study model robustness under this practical set up.

We carry out a usability study to observe how users highlight 
cells in a table.
Based on the imperfect cell selections that users 
produce, we automatically create additional data examples 
by corrupting examples from the original ToTTo dataset. 
We then fine-tune state-of-the-art table-to-text neural generation 
models with this additional data. We compare the performance of
models fine-tuned only with clean cell highlights versus those 
trained with additional noisy cell highlights, both on a test 
set with clean and noisy highlights. 
Experimental results show that models fine-tuned with clean cell
highlights only perform well on clean test cases (i.e., 
performance drops dramatically when evaluated on noisy cell highlights). That is,
these models do not generalise well in practice with user noisy cell selections.
In contrast, the proposed training scheme with additional noisy 
cell highlights not only makes user controlled table-to-text models
achieve better performance in practical scenarios, but 
it also boosts performance on perfect inputs. 
Experimental results show that models fine-tuned 
with our proposed training regime gain 4.85 BLEU 
points on noisy and 1.4 BLEU points on clean highlights; and
achieve comparable state-of-the-art performance on the ToTTo
dataset.\footnote{\href{https://github.com/google-research-datasets/ToTTo}{ToTTo leaderboard.}}

\section{Methodology}

We describe the process for creating user noisy cell highlights 
from examples in ToTTo \cite{parikh2020totto} (\S\ref{sec:user:cells}
and \S\ref{sec:construct}). 
Then, we evaluate models optimized with 
the standard training scheme (i.e., only on clean cell highlights) 
on the created noisy test cases. Results show that these models
perform poorly. To improve model robustness, we propose a new
learning regime described in \S\ref{sec:augment_training_data}.
To further improve performance, we 
fine-tune with Reinforcement Learning (RL) based optimisation
(\S\ref{sec:rl}). Finally, \S\ref{sec:learning} summarises the learning schemes and objective functions we propose for robust
user controlled table-to-text generation.

\subsection{How Do Users Select Cells?} 
\label{sec:user:cells}

To understand how users proceed when exploring a table and 
selecting cells we carry out a human study using examples
from the ToTTo dataset. Participants
are given a plain table (i.e., without highlights) and asked
to highlight cells according to an exploratory intention. 
For a more controlled setting, we give the sentence associated
to the table as the exploratory intention. In this way, we avoid
ambiguous post-selection analysis of what the user intention
was. In addition, this allows us to compare user selections
with reference highlights as well as differences (if any)
in model generated texts given user and reference highlights.

We conduct this study on Amazon Mechanical Turk (the interface is 
described in Appendix~\ref{app:screen}). We collect 90 user highlights 
(3 participants, volunteers known by the authors, and 30 examples
from the validation set) and observe the following noise in 
their highlights. Participants apply different criteria to 
include (or not) table headers; select additional cells in columns/rows 
around cells containing relevant content; and do not
select cells that contain content relevant to the intention.

\subsection{Creating User Noisy Cell Selections}
\label{sec:construct}

Given the input table $T$, the reference text $S$, and the reference 
highlight cells $H\in T$ relevant for generating $S$, 
we create noisy user cell selections as follows.
We provide an example illustrating each noise type in Figure~\ref{tab:examples}.

\paragraph{Noise 1: Additional Table Cells} In practical scenarios, users may accidentally select random cells that are not related to their exploration intention. Thus, we randomly select $k$ cells from the table cells in $T$ that are not in $H$ and add them into $H$ to form a corrupted input $H_1$. 
$H_1$ can be viewed as adding irrelevant information in the generation of the target text $S$.

\paragraph{Noise 2: Table Headers as Additional Inputs} Reference highlight cells  
in the ToTTo dataset do not cover table headers. As we have observed, users may decide to include (or not) table headers in different cases. 
To simulate this, we first retrieve table headers corresponding to highlight cells in $H$. Then, we randomly select $k$ unique headers and add them into $H$ to get the corrupted input $H_2$.

\paragraph{Noise 3: Similar Table Cells} For this type of noise, we select cells that are in the same row/column as the highlight cells. The intuition, as seen in the user study, is that these cells will have similar semantics to those cells underlying the exploratory intention and users tend to select them. For $H_3$, we first retrieve table cells that are in the same row/column as highlight cells. Then, we randomly select $k$ unique cells thereof and add them into $H$.

\paragraph{Noise 4: Remove Cells from $H$} 
Users also miss some of the highlight cells in $H$. 
For this type of noise, we first retrieve those cells in $H$ that are irrelevant (i.e., their content is not expressed in) for generating $S$. 
After getting the irrelevant cells in $H$, we randomly choose $k$ thereof and remove them from $H$ to create $H_4$.

\subsection{Augmenting the Training Dataset}
\label{sec:augment_training_data}
We propose to fine-tune models on the training set augmented 
with noisy data. We extend the original ToTTo training set 
$\mathcal{D} = \{(T, S, H)\}^{|\mathcal{D}|}_{j=1}$ 
with data instances with user noisy cell selections. 
Specifically, we replace data instances with clean cell selections
$H$ in $\mathcal{D}$ with corrupted data instances with noisy cell
selections $H_i$. This results in a training set $\mathcal{D}_i$ 
consisting of noisy cell selections of noise type $i$. 
The final training set $\mathcal{D}_{final}$ contains both clean 
and corrupted data instances, its size is 603,805 (5 times 
the size of the original training set), and 
it is defined as
 $\mathcal{D}_{final} = \mathcal{D} \cup \mathcal{D}_1 \cup \mathcal{D}_2 \cup \mathcal{D}_3 \cup \mathcal{D}_4$.
We set $k=1$ for creating data instances of type Noise 1, Noise 2, 
and Noise 3. This is because the average number of highlight cells in 
ToTTo dataset is small (3.55). To create instances of type Noise 4, 
we remove all irrelevant cells found in $H$.

\subsection{Robustness via Sequence Level Training}
\label{sec:rl}

Inspired by PlanGen \cite{su2021plan}, to further enhance the 
robustness of table-to-text models on clean and noisy cell selections, 
we further fine-tune model parameters with Reinforcement Learning (RL)
\cite{williams1992simple}. 
Formally, given an input data pair $\{T, S, H\} \in \mathcal{D}_{final}$
and a sampled output sequence $S^{'} = (S_1^{'},...,S_{|S^{'}|}^{'})$, 
the RL training objective is formulated as:
\vspace{-1ex}
\begin{equation}
\small
    \mathcal{L_{RL}} = -R(S, S^{'}) \sum^{|S^{'}|}_{i=1} log \, P \, \bigg ( S^{'}_i \, \big| \, S^{'}_{<i}, E (T, H) \bigg )
\end{equation} 
where $E(\cdot)$ denotes the encoder module of a table-to-text generator.
The reward function $R(S, S^{'})$ measures the similarity between the
reference text and the text generated by the model; it is formulated as
$R(S, S^{'}) = B(S, S^{'})$
where $B(\cdot, \cdot)$ is the BLEU score \cite{BLEU}. 
By doing this, we make the outputs of both clean and noisy cell selections
to be more similar to the reference texts. This implicitly improves 
the similarity between outputs of clean and noisy cell selections.

\subsection{Table-to-Text Generation Models}
\label{sec:learning}

Our models are based on BART \cite{bart}. 
We fine-tune them for user controlled table-to-text generation as 
follows. Given a training data pair $\{T, S, H\}$, the 
fine-tuning process proceeds in two stages. The first stage 
fine-tunes the model with a conventional conditional language 
modelling training objective: 
\vspace{-1ex}
\begin{equation}
  \small
    \mathcal{L_{LM}} = - \sum^{|S|}_{i=1} log \, P \, \bigg ( S_i \, \big| \, S_{1:i-1}, E (T,H) \bigg )
\end{equation} where $E$ denotes the encoder of the table-to-text generator. 
The second stage further adjusts model parameters by using
$\mathcal{L}_{mix} = \mathcal{L_{LM}} + \mathcal{L_{RL}}$.

\begin{table}[t]
\vskip 2mm
\begin{center}
\begin{footnotesize}
\begin{tabular}{l@{\hspace{0.09cm}}c@{\hspace{3pt}}c@{\hspace{3pt}}c@{\hspace{3pt}}r@{\hspace{3pt}}}
\hline
\textbf{Model} & Clean   &  Noise & Noise & \#Param\\
               &   & Avg. & Var. & \\
\hline
BART-BASE (clean)     & 47.8  & 44.0 &9.09 &\textbf{141M} \\
BART-LARGE (clean)    & 48.6   &43.9 &14.43 &408M \\
\hline
BART-BASE ($\mathcal{D}_{final}$) &48.5  &48.03 &0.16 &\textbf{141M}       \\
BART-BASE + RL ($\mathcal{D}_{final}$) &49.2   &\textbf{48.85} &0.14 &\textbf{141M}   \\
BART-LARGE ($\mathcal{D}_{final}$)  &49.1 &48.16 &0.69 &408M  \\
BART-LARGE + RL ($\mathcal{D}_{final}$)   &\textbf{49.6}  &48.75 &0.60 &408M  \\
\hline
\end{tabular}
\end{footnotesize}
\caption{BLEU scores on clean and noisy development sets. 
Average BLEU score across the four noisy development sets (Noise Avg.).
Variance of BLEU scores across the four noisy development sets (Noise Var.). 
Model parameters (\#Param). 
The attribute in parenthesis indicates the dataset used for model fine-tuning.
}
\label{tab:clean and noise}
\end{center}
\vskip -3mm
\end{table}

\begin{figure*}[th]
\begin{scriptsize}
\begin{tabular}{c}
\begin{tabular}{|llllll|l|lllll|}
\cline{1-6} \cline{8-12}
\multicolumn{6}{|l|}{Kosuke Matsuura}                                                                                                                                                                                                                                                                                                                                                                                                                                                            &  & \multicolumn{5}{l|}{Asian Beach Games}                                                                                                                                                                                                                                                                                                                                                                                                                                                                                                                                    \\ \cline{1-6} \cline{8-12} 
\multicolumn{6}{|l|}{Section Title: IndyCar Series}                                                                                                                                                                                                                                                                                                                                                                                                                                              &  & \multicolumn{5}{l|}{Section Title: List of Asian Beach Game}                                                                                                                                                                                                                                                                                                                                                                                                                                                                                                              \\ \cline{1-6} \cline{8-12} 
\multicolumn{1}{|l|}{Year}                                              & \multicolumn{1}{l|}{Team}                                                                                                                  & \multicolumn{1}{l|}{14}                                                & \multicolumn{1}{l|}{16}                                                & \multicolumn{1}{l|}{Rank}                                             & Points                                          &  & \multicolumn{1}{l|}{Edition}                                                                  & \multicolumn{1}{l|}{Year}                                                                                         & \multicolumn{1}{l|}{City}                                                                                            & \multicolumn{1}{l|}{Start Date}                                                                                           & End Date                                                                                           \\ \cline{1-6} \cline{8-12} 
\multicolumn{1}{|l|}{2004}                                              & \multicolumn{1}{l|}{\begin{tabular}[c]{@{}l@{}}Super Aguri\\ Fernandez\\ Racing\end{tabular}}                                              & \multicolumn{1}{l|}{CHI Ret}                                           & \multicolumn{1}{l|}{TX2 Ret}                                           & \multicolumn{1}{l|}{14th}                                             & \cellcolor[HTML]{FFCE93}280                     &  & \multicolumn{1}{l|}{IV}                                                                       & \multicolumn{1}{l|}{\cellcolor[HTML]{FFCE93}2014}                                                                 & \multicolumn{1}{l|}{Phuket}                                                                                          & \multicolumn{1}{l|}{14 November}                                                                                          & 23 November                                                                                        \\ \cline{1-6} \cline{8-12} 
\multicolumn{1}{|l|}{\cellcolor[HTML]{FFFC9E}2005}                      & \multicolumn{1}{l|}{\cellcolor[HTML]{FFFC9E}\begin{tabular}[c]{@{}l@{}}Super Aguri\\ Fernandez\\ Racing\end{tabular}}                      & \multicolumn{1}{l|}{\cellcolor[HTML]{FFFC9E}SNM 6}                     & \multicolumn{1}{l|}{\cellcolor[HTML]{FFFC9E}WGL 6}                     & \multicolumn{1}{l|}{\cellcolor[HTML]{FFFC9E}14th}                     & 320                                             &  & \multicolumn{1}{l|}{V}                                                                        & \multicolumn{1}{l|}{\cellcolor[HTML]{FFFC9E}2016}                                                                 & \multicolumn{1}{l|}{\cellcolor[HTML]{FFFC9E}Da Nang}                                                                 & \multicolumn{1}{l|}{\cellcolor[HTML]{FFFC9E}24 September}                                                                 & \cellcolor[HTML]{FFFC9E}3 October                                                                  \\ \cline{1-6} \cline{8-12} 
\multicolumn{1}{|l|}{}                                                  & \multicolumn{1}{l|}{}                                                                                                                      & \multicolumn{1}{l|}{}                                                  & \multicolumn{1}{l|}{}                                                  & \multicolumn{1}{l|}{}                                                 &                                                 &  & \multicolumn{1}{l|}{VI}                                                                       & \multicolumn{1}{l|}{\cellcolor[HTML]{FFFC9E}2020}                                                                 & \multicolumn{1}{l|}{\cellcolor[HTML]{FFFC9E}Sanya}                                                                   & \multicolumn{1}{l|}{\cellcolor[HTML]{FFFC9E}24 November}                                                                  & \cellcolor[HTML]{FFFC9E}5 December                                                                 \\ \cline{1-6} \cline{8-12} 
\multicolumn{6}{|l|}{\begin{tabular}[c]{@{}l@{}}\textbf{Reference}: In 2005, Kosuke Matsuura again drove for Super Aguri \\ Fernandez  Racing, and again finished 14th with a best place finish of 6th \\ 
 in the two races.\\ \\ \textbf{Ours}: In 2005, Kosuke Matsuura drove for Super Aguri Fernandez Racing \\in the IndyCar Series and finished 14th in points.\\  \\ \textbf{Baseline}: In 2005, Kosuke Matsuura drove for Super Aguri Fernandez Rac-\\-ing and finished 14th in the WGL 6 and 280 points.\end{tabular}} &  & \multicolumn{5}{l|}{\begin{tabular}[c]{@{}l@{}}\textbf{Reference}: The last Asian Beach Games was held in Danang, Vietnam\\ from 24 September to 3 October 2016, while the next will be held\\ in 2020 in Sanya, China, the first to breakaway from the 2-year cycle.\\ \\ \textbf{Ours}: The Asian Beach Games are scheduled to be held in Da Nang,\\ Vietnam from September 24 to October 3, 2016 and in Sanya, China \\ in 2020.\\ \\ \textbf{Baseline}: The Asian Beach Games were held from 2014 to 2016 in Da\\ Nang, Vietnam and from 3 October to 3 October 2020 in Sanya, China.\end{tabular}} \\ \cline{1-6} \cline{8-12} 
\end{tabular}
\\
\\
\begin{tabular}{lllllllll}
\cline{1-3} \cline{5-9}
\multicolumn{3}{|l|}{List of rulers of Brittany}                                                                                                                                            & \multicolumn{1}{l|}{} & \multicolumn{5}{l|}{Iain Glein}                                                                                                                                                                                                                                                        \\ \cline{1-3} \cline{5-9} 
\multicolumn{3}{|l|}{Section Title : House of Montfort}                                                                                                                                                                                                                                                                                                                                                                  & \multicolumn{1}{l|}{} & \multicolumn{5}{l|}{Section Title: Awards and nominations}                                                                                                                                                                                                                             \\ \cline{1-3} \cline{5-9} 
\multicolumn{1}{|c|}{\cellcolor[HTML]{FFCE93}Name}                                                                                      & \multicolumn{1}{c|}{Birth}                                                                                                                            & \multicolumn{1}{c|}{Death}                                                                                             & \multicolumn{1}{l|}{} & \multicolumn{1}{l|}{Year}                         & \multicolumn{1}{l|}{Title}                                 & \multicolumn{1}{l|}{Award}                               & \multicolumn{1}{l|}{Category}                           & \multicolumn{1}{l|}{Result}                      \\ \cline{1-3} \cline{5-9} 
\multicolumn{1}{|c|}{\cellcolor[HTML]{FFFC9E}\begin{tabular}[c]{@{}c@{}}Peter II the Simple\\ (Pêr II)\\ 1450–1457\end{tabular}}        & \multicolumn{1}{c|}{\begin{tabular}[c]{@{}c@{}}7 July 1418 \end{tabular}}                             & \multicolumn{1}{c|}{\begin{tabular}[c]{@{}c@{}}22 September 1457\\ Nantes aged 41\end{tabular}}                        & \multicolumn{1}{l|}{} & \multicolumn{1}{l|}{\cellcolor[HTML]{FFFC9E}1990} & \multicolumn{1}{l|}{\cellcolor[HTML]{FFFC9E}Silent Scream} & \multicolumn{1}{l|}{\cellcolor[HTML]{FFFC9E}Silver Bear} & \multicolumn{1}{l|}{\cellcolor[HTML]{FFFC9E}Best Actor} & \multicolumn{1}{l|}{\cellcolor[HTML]{FFCE93}Won} \\ \cline{1-3} \cline{5-9} 
\multicolumn{1}{|c|}{\cellcolor[HTML]{FFFC9E}\begin{tabular}[c]{@{}c@{}}Arthur III the Justicier\\ (Arzhur III) 1457–1458\end{tabular}} & \multicolumn{1}{c|}{\begin{tabular}[c]{@{}c@{}}24 August 1393\end{tabular}} & \multicolumn{1}{c|}{\cellcolor[HTML]{FFFC9E}\begin{tabular}[c]{@{}c@{}}26 December 1458\\ Nantes aged 65\end{tabular}} & \multicolumn{1}{l|}{} & \multicolumn{5}{l|}{}                                                                                                                                                                                                                                                                  \\ \cline{1-3} \cline{5-9} 
\multicolumn{3}{|l|}{\begin{tabular}[c]{@{}l@{}}\textbf{Reference}: At the very end of his life, Arthur III became duke of Brittany, \\succeeding Peter II. \\ \\ \textbf{Ours}: Arthur III the Justicier was Duke of Brittany from 1457 until his death\\ in 1458, succeeding Peter II the Simple.\\ \\ \textbf{Baseline}: Arthur III (26 December 1458) was Duke of Brittany from 1450 \\ to his death.\end{tabular}}                                                                                                                                            & \multicolumn{1}{l|}{} & \multicolumn{5}{l|}{\begin{tabular}[c]{@{}l@{}}\textbf{Reference}: In 1990, Glen won the Silver Bear for the Best Actor in the \\ Silent Scream.\\ \\ \textbf{Ours}: In 1990, Iain Glen won the Silver Bear for Best Actor for Silent \\ Scream.\\ \\ \textbf{Baseline}: In 1990, Iain Glen received the Silver Bear for Best Actor for\\ Silent Scream.\end{tabular}}                                              \\ \cline{1-3} \cline{5-9} 
                                              
\end{tabular}  
\end{tabular}
\end{scriptsize}
\vspace{-1ex}
\caption{Model outputs for synthetic noisy cell selections of type Noise 1 (left top) and Noise 2 (left bottom), and for user noisy cell selections from the human study of type Noise 3 (right top) and Noise 4 (right bottom) .}
\label{tab:examples}
\end{figure*}

\section{Experimental Results}
\label{sec:experimental_results}

Implementation details for our table-to-text generation
models can be found in Appendix~\ref{app:impl:details}. 
We use the same hyperparameters as the baseline in the ToTTo \cite{parikh2020totto}.

As shown in Table~\ref{tab:clean and noise} (detailed results per Noise type
are given in Appendix~\ref{app:detailed:results}), when using the 
training scheme with clean cell highlights, the average BLEU score 
of \textbf{BART-BASE (clean)} drops from 47.8 to 44 when tested on 
noisy cell selections. Similar trend can be seen for \textbf{BART-LARGE
(clean)} with a BLUE score drop from 48.6 to 43.9. In addition, the 
``Noise Variance'' of \textbf{BART-BASE (clean)} and \textbf{BART-LARGE
(clean)} is large, indicating that these models are not stable (or robust) 
to different types of noisy cell selections. 
All this suggests that a training scheme with carefully selected 
cells alone results in systems that perform poorly in practical 
scenarios with user interactions.

\begin{table}[t]
\centering
\begin{footnotesize}

\begin{tabular}{ll@{\hspace{6pt}}c@{\hspace{4pt}}c@{\hspace{4pt}}c@{\hspace{4pt}}}
\hline
  & \textbf{Model}       & \textbf{FL} & \textbf{FA} & \textbf{CC}  \\ \hline
\multicolumn{1}{c}{\multirow{2}{*}{clean}} & BART-LARGE (clean) & 0.83 & 0.83  & 0.89           \\
\multicolumn{1}{c}{}                       & BART-LARGE + RL ($\mathcal{D}_{final}$) & 0.88  & 0.89   & 0.93   \\
\hline
\multirow{2}{*}{Noisy}                     & BART-LARGE (clean) & 0.80   & 0.81 & 0.87      \\
                                           & BART-LARGE + RL ($\mathcal{D}_{final}$)   & 0.89 & 0.91 & 0.91   \\ \hline
\end{tabular}
\vspace{-2mm}
    \caption{Results of Human Evaluation. Percentage of outputs perceived as Fluent (FL), Faithful (FA), and better Covering selected Cells (CC).}
\label{tab:human eval}
\end{footnotesize}
\vskip -3.5mm
\end{table}

\begin{table}[t]  
  \centering  
  {\footnotesize
    \begin{tabular}{cccc}  
    \hline
    \multirow{2}{*}{Method}&  
    \multicolumn{3}{c}{ Overall}\\  
    &BLEU&PARENT&BLEURT \\ 
    \hline
    NCP & 19.2 & 29.2 & -0.576 \\  
    Pointer Generator &41.6 &51.6 & 0.076 \\  
    Bert-to-Bert & 44.0 & 52.6 & 0.121 \\  
    LATTICE & 48.4 & 58.1 &  0.222 \\ 
    T5-3B & \textbf{49.5} & 58.4 & 0.230 \\  
    PlanGen & 49.2 & 58.7 & \textbf{0.249}  \\  
    Ours & 49.3 & \textbf{58.8} & 0.235 \\  
    \hline
    \end{tabular}
    }
    \vspace{-2mm}
    \caption{ToTTo test set results. All reported results can be found in the ToTTo leaderboard. }
    \label{tab:leaderboard-part}  
\vskip -3.5mm
\end{table}

In contrast, we observe that our proposed learning scheme makes 
generators achieve better performance both on clean and noisy cell
selections. On clean cell selections (ToTTo original development set), 
the model trained using the proposed learning scheme 
\textbf{BART-BASE ($\mathcal{D}_{final}$)} outperforms the model
using the same pre-trained model but fine-tuned with the standard 
learning scheme \textbf{BART-BASE (clean)} by 0.7 BLEU scores. 
On noisy cell selections, \textbf{BART-BASE ($\mathcal{D}_{final}$)} 
outperforms \textbf{BART-BASE (clean)} by 4.03 BLEU points on average. 
In addition, \textbf{BART-BASE ($\mathcal{D}_{final}$)} 
has a small ``Noise Variance'' score across four noisy and one clean development sets, suggesting that the proposed learning scheme 
can make controlled table-to-text generators more robust and less 
sensitive to various types of noisy cell selections. Fine-tuning with 
RL, \textbf{BART-BASE + RL ($\mathcal{D}_{final}$)}, can further boost models' performance. 

In Appendix~\ref{app:detailed:results} we provide additional experiments on ablation results 
on the contribution of each Noise dataset, training with a subset 
of $\mathcal{D}_{final}$ (i.e., training with one fifth of the data also improves robustness), and evaluating on cases with different amount of noise (i.e., our approach 
generalises better to cases with higher values of $k$).

To gain insights on how the improvements are perceived in generated
descriptions, we conduct a human evaluation. We follow the setup 
described in \cite{parikh2020totto}.
We sample 100 development instances and have five human judges 
(voluntary MSc level students fluent in English) to
annotate them across three criteria. 
\textbf{Fluency} (users select amongst {\small \textit{Fluent}}, 
{\small \textit{Mostly Fluent}}, and {\small \textit{Not Fluent}}; we 
report the percentage of outputs annotated as {\small \textit{Fluent}};
\textbf{Faithfulness} (a candidate sentence is considered to be faithful if all the information in it is supported by the highlight cells and metadata of the table; we report the percentage of outputs that users annotate
as faithful); and \textbf{Covered Cells} (the percentage of highlighted
cells that the candidate sentence covers; we report average percentage of covered cells across all sampled instances).
Table~\ref{tab:human eval} shows that judges find outputs by the 
model variants fine-tuned with the proposed regime more faithful, 
fluent and with better cell coverage.

We choose the best performing model, \textbf{BART-LARGE + RL ($\mathcal{D}_{final}$)}, fine-tuned with the proposed approach and 
compare it with state-of-the-art models on 
the ToTTo test set. 
These are NCP \cite{puduppully2019data}, Pointer-Generator \cite{see2017get}, Bert-to-Bert 
\cite{parikh2020totto}, and T5-3B \cite{2020t5},  LATTICE \cite{wang2022robust}, and PlanGen \cite{su2021plan}.
Table~\ref{tab:leaderboard-part} shows overall results
(detailed overlap/non-overlap results are provided in 
Appendix~\ref{app:detailed:results}). Our model performs in 
par with T5-3B and PlanGen despite the fact that the first one 
has more parameters and the second one posses a dedicated planning step.

Figure~\ref{tab:examples} shows two instances of synthetic noisy cell
selections of type Noise 1 (i.e., accidentally selected random cell not related to the exploration intention) and type Noise 2 (i.e., random criteria for header
selection); and two instances of user noisy cell selection from the human study of type Noise 3 (i.e., highlight {\small \textit{2014}} semantically close to cells in the exploratory intention) and Noise 4 (i.e.,
{\small \textit{won}} is not highlighted). Cells in yellow indicate original highlights from the
ToTTo dataset and those in orange are noisy selections.
In both cases, the outputs produced by the model fine-tuned 
with the proposed regime are not affected by noise and 
show better coverage, factual accuracy, and lexicalisation. This illustrates human evaluation preferences.

\section{Conclusion}

We study the performance of user controlled table-to-text generation.
We show that standard training schemes with only carefully selected 
cells causes poor robustness of generators in practice when confronted 
with user noisy cell selections. To address this, 
we introduce a training scheme with simulated user noisy cell selections.
Experimental results show that generators optimized with our proposed
scheme can achieve better performance on both clean and noisy cell selections. In the future, it would be interesting to investigate how to apply our
approach to other data-to-text datasets to improve model generalisation.

\section{Acknowledgments}

We thank the anonymous reviewers for their feedback.  We gratefully acknowledge the support of the UK Engineering and Physical Sciences Research Council (award number 681760).

\section*{Limitations}

We create synthetic data simulating real users interactions (i.e.,
user cell selections on a table).  However, the automatic noise generation 
method does not cover all possible user interactions and may fail
to exactly reproduce them in some cases. 
For example, our process for creating Noise 3 randomly highlights cells 
in the same row/column as a reference highlighted cell. However, 
the probability distribution of a user highlighting a cell around a 
reference highlighted cell is not always uniform, but in some cases based on 
some reasoning process about the concerned cells. 
In the future, it would be interesting to investigate how to simulate 
this reasoning process to predict where the user is likely to 
highlight cells.
Nevertheless, the set of noise types that we propose in this work shows 
that models trained only on cleaned data are brittle.

\bibliography{anthology,custom}

\begin{thebibliography}{16}
\expandafter\ifx\csname natexlab\endcsname\relax\def\natexlab#1{#1}\fi

\bibitem[{Kale and Rastogi(2020)}]{kalerastogi-2020-text}
Mihir Kale and Abhinav Rastogi. 2020.
\newblock Text-to-text pre-training for data-to-text tasks.
\newblock In \emph{Proceedings of the 13th International Conference on Natural
  Language Generation}, pages 97--102.

\bibitem[{Kingma and Ba(2014)}]{kingma2014adam}
Diederik~P Kingma and Jimmy Ba. 2014.
\newblock Adam: A method for stochastic optimization.
\newblock \emph{arXiv preprint arXiv:1412.6980}.

\bibitem[{Lebret et~al.(2016)Lebret, Grangier, and
  Auli}]{lebret-etal-2016neural}
R{\'e}mi Lebret, David Grangier, and Michael Auli. 2016.
\newblock Neural text generation from structured data with application to the
  biography domain.
\newblock In \emph{Proceedings of the 2016 Conference on Empirical Methods in
  Natural Language Processing}, pages 1203--1213.

\bibitem[{Lewis et~al.(2020)Lewis, Liu, Goyal, Ghazvininejad, Mohamed, Levy,
  Stoyanov, and Zettlemoyer}]{bart}
Mike Lewis, Yinhan Liu, Naman Goyal, Marjan Ghazvininejad, Abdelrahman Mohamed,
  Omer Levy, Veselin Stoyanov, and Luke Zettlemoyer. 2020.
\newblock Bart: Denoising sequence-to-sequence pre-training for natural
  language generation, translation, and comprehension.
\newblock In \emph{Proceedings of the 58th Annual Meeting of the Association
  for Computational Linguistics}, pages 7871--7880.

\bibitem[{Mille et~al.(2021)Mille, Dhole, Mahamood, Perez-Beltrachini, Gangal,
  Kale, van Miltenburg, and Gehrmann}]{gem-sets-2021}
Simon Mille, Kaustubh Dhole, Saad Mahamood, Laura Perez-Beltrachini, Varun
  Gangal, Mihir Kale, Emiel van Miltenburg, and Sebastian Gehrmann. 2021.
\newblock {Automatic Construction of Evaluation Suites for Natural Language
  Generation Datasets}.
\newblock In \emph{Thirty-fifth Conference on Neural Information Processing
  Systems Datasets and Benchmarks Track}.
\newblock (NeurIPS 2021).

\bibitem[{Papineni et~al.(2002)Papineni, Roukos, Ward, and Zhu}]{BLEU}
Kishore Papineni, Salim Roukos, Todd Ward, and Wei-Jing Zhu. 2002.
\newblock Bleu: a method for automatic evaluation of machine translation.
\newblock In \emph{Proceedings of the 40th annual meeting of the Association
  for Computational Linguistics}, pages 311--318.

\bibitem[{Parikh et~al.(2020)Parikh, Wang, Gehrmann, Faruqui, Dhingra, Yang,
  and Das}]{parikh2020totto}
Ankur Parikh, Xuezhi Wang, Sebastian Gehrmann, Manaal Faruqui, Bhuwan Dhingra,
  Diyi Yang, and Dipanjan Das. 2020.
\newblock Totto: A controlled table-to-text generation dataset.
\newblock In \emph{Proceedings of the 2020 Conference on Empirical Methods in
  Natural Language Processing (EMNLP)}, pages 1173--1186.

\bibitem[{Perez-Beltrachini and Lapata(2018)}]{perez-lapata2018}
Laura Perez-Beltrachini and Mirella Lapata. 2018.
\newblock Bootstrapping generators from noisy data.
\newblock In \emph{North American Chapter of the Association for Computational
  Linguistics}, New Orleans, Louisiana. Association for Computational
  Linguistics.
\newblock (NAACL 2018).

\bibitem[{Puduppully et~al.(2019)Puduppully, Dong, and
  Lapata}]{puduppully2019data}
Ratish Puduppully, Li~Dong, and Mirella Lapata. 2019.
\newblock Data-to-text generation with content selection and planning.
\newblock In \emph{Proceedings of the AAAI conference on artificial
  intelligence}, volume~33, pages 6908--6915.

\bibitem[{Raffel et~al.(2020)Raffel, Shazeer, Roberts, Lee, Narang, Matena,
  Zhou, Li, and Liu}]{2020t5}
Colin Raffel, Noam Shazeer, Adam Roberts, Katherine Lee, Sharan Narang, Michael
  Matena, Yanqi Zhou, Wei Li, and Peter~J. Liu. 2020.
\newblock Exploring the limits of transfer learning with a unified text-to-text
  transformer.
\newblock \emph{Journal of Machine Learning Research}, 21(140):1--67.

\bibitem[{See et~al.(2017)See, Liu, and Manning}]{see2017get}
Abigail See, Peter~J Liu, and Christopher~D Manning. 2017.
\newblock Get to the point: Summarization with pointer-generator networks.
\newblock \emph{arXiv preprint arXiv:1704.04368}.

\bibitem[{Su et~al.(2021)Su, Vandyke, Wang, Fang, and Collier}]{su2021plan}
Yixuan Su, David Vandyke, Sihui Wang, Yimai Fang, and Nigel Collier. 2021.
\newblock Plan-then-generate: Controlled data-to-text generation via planning.
\newblock In \emph{Findings of the Association for Computational Linguistics:
  EMNLP 2021}, pages 895--909.

\bibitem[{Wang et~al.(2022)Wang, Xu, Szekely, and Chen}]{wang2022robust}
Fei Wang, Zhewei Xu, Pedro Szekely, and Muhao Chen. 2022.
\newblock Robust (controlled) table-to-text generation with structure-aware
  equivariance learning.
\newblock \emph{arXiv preprint arXiv:2205.03972}.

\bibitem[{Williams(1992)}]{williams1992simple}
Ronald~J Williams. 1992.
\newblock Simple statistical gradient-following algorithms for connectionist
  reinforcement learning.
\newblock \emph{Machine learning}, 8(3):229--256.

\bibitem[{Wiseman et~al.(2018)Wiseman, Shieber, and
  Rush}]{wiseman-etal-2018-learning}
Sam Wiseman, Stuart Shieber, and Alexander Rush. 2018.
\newblock \href {https://doi.org/10.18653/v1/D18-1356} {Learning neural
  templates for text generation}.
\newblock In \emph{Proceedings of the 2018 Conference on Empirical Methods in
  Natural Language Processing}, pages 3174--3187, Brussels, Belgium.
  Association for Computational Linguistics.

\bibitem[{Wolf et~al.(2020)Wolf, Debut, Sanh, Chaumond, Delangue, Moi, Cistac,
  Rault, Louf, Funtowicz, Davison, Shleifer, von Platen, Ma, Jernite, Plu, Xu,
  Le~Scao, Gugger, Drame, Lhoest, and Rush}]{wolfetal-2020-transformers}
Thomas Wolf, Lysandre Debut, Victor Sanh, Julien Chaumond, Clement Delangue,
  Anthony Moi, Pierric Cistac, Tim Rault, Remi Louf, Morgan Funtowicz, Joe
  Davison, Sam Shleifer, Patrick von Platen, Clara Ma, Yacine Jernite, Julien
  Plu, Canwen Xu, Teven Le~Scao, Sylvain Gugger, Mariama Drame, Quentin Lhoest,
  and Alexander Rush. 2020.
\newblock Transformers: State-of-the-art natural language processing.
\newblock In \emph{Proceedings of the 2020 Conference on Empirical Methods in
  Natural Language Processing: System Demonstrations}, pages 38--45.

\end{thebibliography}
\bibliographystyle{acl_natbib}

\appendix

\section{Detailed and Ablation Results}
\label{app:detailed:results}

Table~\ref{tab:clean and noise full} provides detailed results for
the different model variants \textbf{(clean)} and \textbf{($\mathcal{D}_{final}$)} evaluated on different development 
sets with different types of noise (cf., Table~\ref{tab:clean and noise} in Section~\ref{sec:experimental_results}).
Table~\ref{tab:leaderboard} provides detailed results comparing 
our model \textbf{BART-LARGE + RL ($\mathcal{D}_{final}$)}  with other state-of-the-art methods in ToTTo's leaderboard (cf., Table~\ref{tab:leaderboard-part} in Section~\ref{sec:experimental_results}).

\begin{table*}[h]
\vskip 2mm
\begin{center}
\begin{small}
\begin{tabular}{lcccccccr}
\hline
\textbf{Model} & Clean   & Noise1  & Noise2&Noise3&Noise4& Noise & Noise & \#Param\\
               & Dev set & Dev set & Dev set & Dev set & Dev set & Average & Variance & \\
\hline
BART-BASE (clean)     & 47.8  &40.6 &45.6 &42.5 &47.3 &44 &9.087 &\textbf{141M} \\
BART-LARGE (clean)    &48.6   &39.8 &46.1 &41.7 &48 &43.9 &14.433 &408M \\
\hline
BART-BASE ($\mathcal{D}_{final}$) &48.5 &47.7 &48.6 &47.9 &47.9 &48.025 &0.156 &\textbf{141M}       \\
BART-BASE + RL ($\mathcal{D}_{final}$) &49.2 &\textbf{48.6} &49.4 &\textbf{48.8} &48.6 &\textbf{48.850} &0.143 &\textbf{141M}   \\
BART-LARGE ($\mathcal{D}_{final}$)  &49.1 &46.9 &48.6 &47.6 &48.6 &48.16 &0.689 &408M  \\
BART-LARGE + RL ($\mathcal{D}_{final}$)   &\textbf{49.6} &47.9 &\textbf{49.7} &48.4 &\textbf{49.0} &48.75 &0.603 &408M  \\
\hline
\end{tabular}
\end{small}
\caption{
BLEU scores of models on clean and noisy ToTTo development set. Average BLEU score across the four noisy development sets (Noise Avg.).
Variance of BLEU scores across the four noisy development sets (Noise Var.).  \#Param denotes the total number of parameters in the model. The attribute in parenthesis indicates the training data we use for training the model. For (clean), models are trained on clean ToTTo training set (i.e. using $\mathcal{D}$). For ($\mathcal{D}_{final}$), the noise-augmented training set described in section \ref{sec:augment_training_data} is applied. For '+RL', the Reinforcement Learning algorithm described in section \ref{sec:rl} is applied. }
\label{tab:clean and noise full}
\end{center}
\vskip -3mm
\end{table*}

\begin{table*}[h]  
  \centering  
  \fontsize{9}{9}\selectfont  
  \begin{threeparttable}  
    \begin{tabular}{ccccccccccc}  
    \toprule  
    \multirow{2}{*}{Method}&  
    \multicolumn{3}{c}{ Overall}&\multicolumn{3}{c}{ Overlap}&\multicolumn{3}{c}{ non-Overlap}\cr  
    \cmidrule(lr){2-4} \cmidrule(lr){5-7}  \cmidrule(lr){8-10} 
    &BLEU&PARENT&BLEURT&BLEU&PARENT&BLEURT&BLEU&PARENT&BLEURT\cr  
    \midrule  
    NCP&19.2&29.2&-0.576&24.5&32.5&-0.491&13.9&25.8&-0.662\cr  
    Pointer Generator&41.6&51.6&0.076&50.6&58.0&0.244&32.2&45.2&-0.092\cr  
    Bert-to-Bert&44.0&52.6&0.121&52.7&58.4&0.259&35.1&46.8&-0.017\cr  
    T5-3B&\textbf{49.5}&58.4&0.230&\textbf{57.5}&62.6&0.351&41.4&54.2&0.108\cr  
    PlanGen&49.2&58.7&\textbf{0.249}&56.9&62.8&\textbf{0.371}&\textbf{41.5}&\textbf{54.6}&\textbf{0.126}\cr  
    Ours&49.3&\textbf{58.8}&0.235&57.1&\textbf{63.4}&0.358&\textbf{41.5}&54.1&0.112\cr  
    \bottomrule  
    
    \end{tabular}  
    \end{threeparttable}  
    \caption{ToTTo test set results. All reported results can be found in the ToTTo leaderboard.}
    \label{tab:leaderboard}  
\end{table*}

\begin{table}[]
\renewcommand\tabcolsep{2pt}
\begin{center}
\begin{scriptsize}
\begin{tabular}{lccccccc}
\hline
Training  & Clean & Noise1 & Noise2 & Noise3 & Noise4 & Noise & Noise \\
Data      & Dev   & Dev    & Dev    & Dev    & Dev  & Avg & Var \\
\hline
$\mathcal{D}_{final}$     & 48.5  & \textbf{47.7} &\textbf{48.6} &\textbf{47.9} &47.9 &\textbf{48.025} &0.156\\
$\mathcal{D}_{final}-\mathcal{D}_1$  & 48.5  & 47.3 &48.4 &47.6 &47.8 &47.775 &0.216 \\
$\mathcal{D}_{final}-\mathcal{D}_2$  & \textbf{48.6} & 47.6 & 48.5 &\textbf{47.9} &\textbf{48.1} & \textbf{48.025} & \textbf{0.143} \\
$\mathcal{D}_{final}-\mathcal{D}_3$   & 48.5 &47.6 &\textbf{48.6} &47.8 &47.9 &47.975 &0.189\\
$\mathcal{D}_{final}-\mathcal{D}_4$   &48.3 &\textbf{47.7} &\textbf{48.6} &\textbf{47.9} &47.4 &47.900 &0.260\\
\hline
\end{tabular}
\end{scriptsize}
\caption{BLEU scores for \textbf{BART-BASE} trained on different training data and evaluated on different development sets. Noise Avg denotes the average BLEU scores on all noisy development sets. Noise Var denotes the variance of BLEU scores on noisy development sets.}
\label{tab:ablation study}
\end{center}
\vskip -3mm
\end{table}

We conduct an ablation study to investigate the impact of 
each type of noise in $\mathcal{D}_{final}$ (see Section~\ref{sec:construct}).
Specifically, we remove 
one of the four noise types at a time from $\mathcal{D}_{final}$, then 
train the \textbf{BART-BASE} model using the remaining data. 
This study shows that all types of user noisy cell selections help to improve performance and robustness (Table~\ref{tab:ablation study}).

We construct corrupted ToTTo development datasets with \emph{different amount of noise} (i.e., different number $k$ of noisy cells) added to each original input highlighted cells. In the ToTTo dataset, there
are on average 3.5 highlighted cells for each table;
when $k = 3$, the injected noise has roughly the same proportion
as the original highlight cells. We then examine
BLEU scores for \textbf{BART-BASE} trained with our
approach and the baseline on these noisy
development sets. 
As shown in Table~\ref{tab:proportion}, performance drops significantly as
more noise is injected, from 47.8 when $k = 0$ (clean) to 34.8 when $k = 3$, for the model trained only on clean cell selections, \textbf{BART-BASE} (clean). It also indicates that the
models trained with our proposed method, \textbf{BART-BASE ($\mathcal{D}_{final}$)} and \textbf{BART-BASE + RL ($\mathcal{D}_{final}$)}, can reduce this performance drop.

\begin{table}[]
\renewcommand\tabcolsep{2pt}
\begin{center}
\footnotesize
\begin{tabular}{lcccc}
\hline
Model        & clean & $k=1$ & $k=2$ & $k=3$ \\
\hline
BART-BASE (clean)                & 47.8 & 42.7  & 38.1  & 34.8  \\
BART-BASE ($\mathcal{D}_{final}$)& 48.5 & 48.1  & 45.9  & 42.3 \\
BART-BASE + RL ($\mathcal{D}_{final}$) & \textbf{49.2} &\textbf{48.8} &\textbf{46.4} &\textbf{42.9} \\
\hline
\end{tabular}
\caption{BLEU scores on input cell highlights with different amounts of noise (development set). $k$ denotes the amount of noise added to the original data point (higher $k$ means more noisy cell highlights are added).}
\label{tab:proportion}
\end{center}
\vskip -3mm
\end{table}

We also combine all noise types with clean data for training in a way that the resulting
dataset has the same size as the original clean dataset. Specifically, we randomly divide the original dataset into five equal parts and replace four of them each by a different type of
noisy data subset; one of the parts is not replaced (i.e., one part of the original clean set is kept). 
We merge these five parts together and call this the mixed dataset $\mathcal{D}_{mix}$. Results in Table~\ref{tab:fix_size} 
indicate that training the model on a substantially smaller subset of clean and noisy data (i.e., a subset of $\mathcal{D}_{final}$) still yields comparable performance on clean data and significant better performance on noisy data.

\begin{table}[]
\begin{footnotesize}
\begin{center}
\begin{tabular}{@{}c|ccc@{}}
\toprule
Dev/Train  & Clean & $\mathcal{D}_{mix}$ & $\mathcal{D}_{final}$  \\ \midrule
Clean      & 47.8  & 47.3 & 48.50 \\
Noise Avg. & 44.0  & 46.8 & 48.03\\ \bottomrule
\end{tabular}
\end{center}
\end{footnotesize}
\caption{BLEU scores of \textbf{BART-BASE} trained on the original dataset, the noise augmented dataset ($\mathcal{D}_{final}$), and a smaller dataset ($\mathcal{D}_{mix})$. Evaluation is on clean and Noise
development sets.}
\label{tab:fix_size}
\end{table}

\section{Implementation Details}
\label{app:impl:details}

The examined models are based on the Huggingface Library \cite{wolfetal-2020-transformers} 
with default model hyperparameters provided by the Library. 
We fine-tune BART \cite{bart} using the proposed learning 
scheme. We use the Adam \cite{kingma2014adam} optimizer, with a learning rate of $2e^{-5}$ and a batch of size 32. We fine-tune with the $\mathcal{L_{LM}}$ objective for 100k steps and $\mathcal{L}_{mix}$ for 50k steps.  

\section{Human Study Interface}
\label{app:screen}

\begin{figure*}[]
\begin{center}
\centerline{\includegraphics[width=150mm]{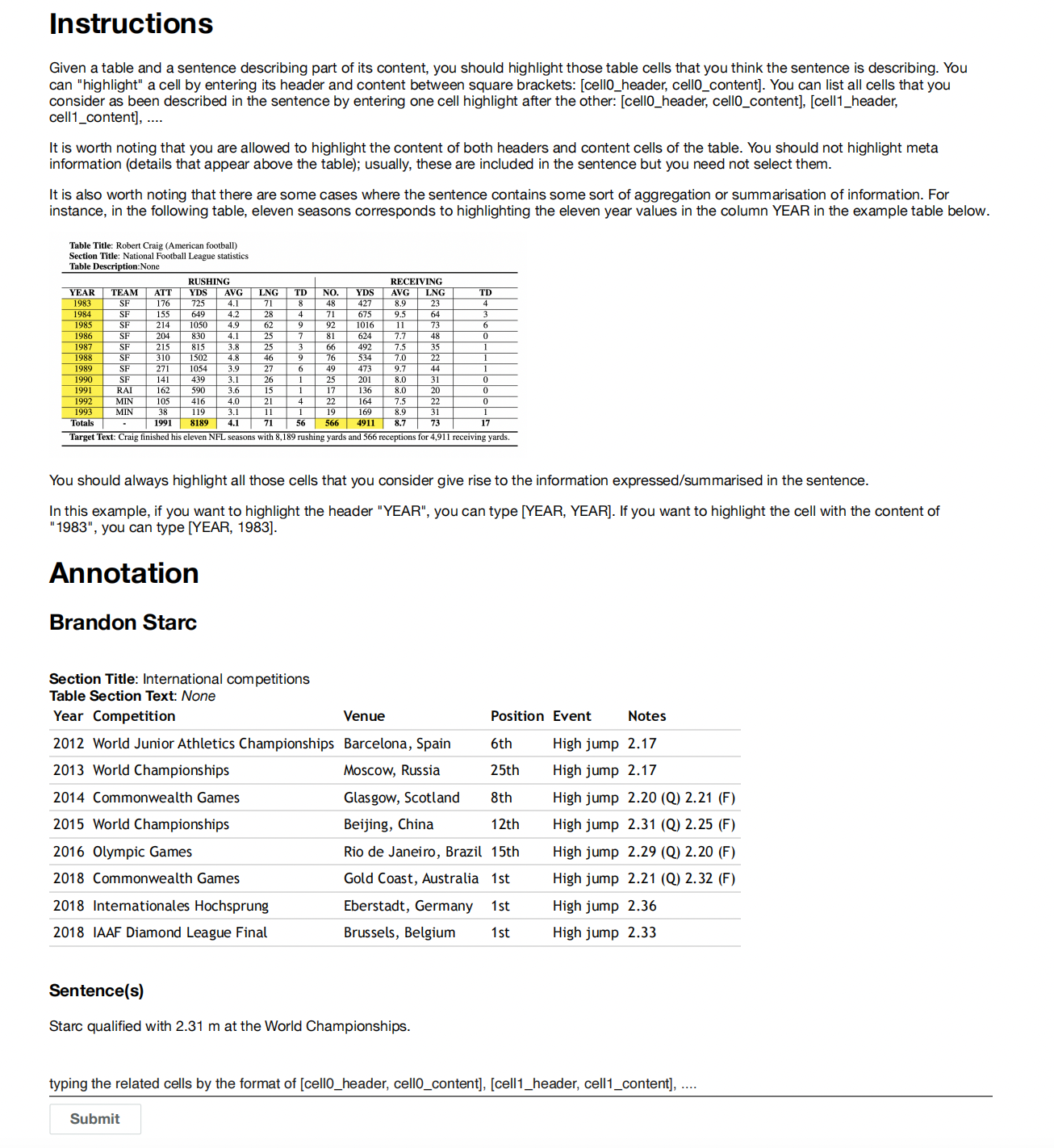}}
\caption{The Amazon Mechanical Turk interface, instructions and annotation form, we use for the human study described in Section~\ref{sec:user:cells}.}
\label{fig:screen}
\end{center}
\vskip -10mm
\end{figure*}

Figure~\ref{fig:screen} shows the Amazon Mechanical Turk interface, instructions and annotation form,
we use for the human study described in Section~\ref{sec:user:cells}. 

\end{document}